\pdfoutput=1
\documentclass[runningheads]{llncs}
\usepackage{amsmath}
\usepackage{booktabs}
\usepackage{CJKutf8}
\usepackage{latexsym}

\usepackage[%
  square,        
  comma,         
  numbers,       
  sort&compress, 
]{natbib}

\usepackage{latexsym}
\usepackage{graphicx}
\usepackage{mathrsfs}
\usepackage{multirow}
\usepackage{color}
\usepackage{tikz-dependency}
\usepackage{times}
\usepackage{url}
\usepackage{caption}
\usepackage{changepage}

\newlength{\offsetpage}
\newenvironment{widepage}{\begin{adjustwidth}{-\offsetpage}{-\offsetpage}%
    \addtolength{\textwidth}{2\offsetpage}}%
{\end{adjustwidth}}
\newlength{\semioffsetpage}
\newenvironment{semiwidepage}{\begin{adjustwidth}{-\semioffsetpage}{-\semioffsetpage}%
  \addtolength{\textwidth}{2\semioffsetpage}}%
{\end{adjustwidth}}
\renewcommand\citet[1]{\citeauthor{#1} (\citeyear{#1}) \cite{#1}}

\begin{document}
\begin{CJK*}{UTF8}{gkai}

\title{Is POS Tagging Necessary or Even Helpful for Neural Dependency Parsing?}

\author{Houquan Zhou\thanks{Houquan Zhou and Yu Zhang make equal contributions to this work. Zhenghua Li is the corresponding author. This work was supported by National Natural Science Foundation of China (Grant No. 61525205, 61876116) and a Project Funded by the Priority Academic Program Development (PAPD) of Jiangsu Higher Education Institutions.} \and
Yu Zhang$^{\star}$ \and
Zhenghua Li \and 
Min Zhang}
\authorrunning{H. Zhou et al.}
%
\institute{Institute of Artificial Intelligence, School of Computer Science and Technology, \\
Soochow University, Suzhou, China \\
\email{hqzhou@stu.suda.edu.cn, yzhang.cs@foxmail.com, \{zhli13,minzhang\}@suda.edu.cn}}
\maketitle

\begin{abstract}

In the pre deep learning era, part-of-speech tags have been considered as indispensable ingredients for feature engineering in dependency parsing.
But quite a few works focus on joint tagging and parsing models to avoid error propagation.
In contrast, recent studies suggest that POS tagging becomes much less important or even useless for neural parsing, especially when using character-based word representations.
Yet there are not enough investigations focusing on this issue, both empirically and linguistically.
To answer this, we design and compare three typical multi-task learning framework, i.e., \textit{Share-Loose}, \textit{Share-Tight}, and \textit{Stack}, for joint tagging and parsing based on the state-of-the-art biaffine parser.
Considering that it is much cheaper to annotate POS tags than parse trees, we also investigate the utilization of large-scale heterogeneous POS tag data.
We conduct experiments on both English and Chinese datasets, and the results clearly show that POS tagging (both homogeneous and heterogeneous) can still significantly improve parsing performance when using the \textit{Stack} joint framework.
We conduct detailed analysis and gain more insights from the linguistic aspect.

\end{abstract}
\section{Introduction}
\begin{figure}[tb]
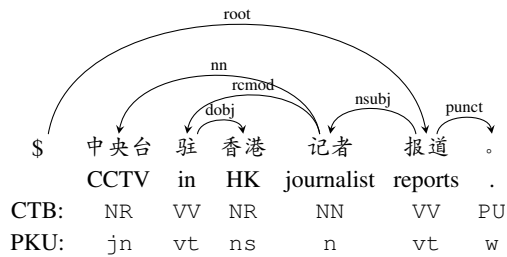

\centering
\begin{dependency}[arc edge, arc angle=80, text only label, label style={above}]
\begin{deptext} [row sep=0.025cm, column sep=0.1cm]
\$ \& 中央台 \& 驻  \& 香港 \& 记者 \& 报道 \&。 \\
\& CCTV \& in  \& HK \& journalist \& reports \&.  \\
CTB: \& \texttt{NR} \& \texttt{VV} \& \texttt{NR} \& \texttt{NN} \& \texttt{VV} \& \texttt{PU} \\
PKU: \& \texttt{jn} \& \texttt{vt} \& \texttt{ns} \& \texttt{n}  \& \texttt{vt} \& \texttt{w}  \\
\end{deptext}
\depedge[edge style={black}]{5}{2}{\color{black}nn}
\depedge[edge style={black}]{5}{3}{\color{black}rcmod}
\depedge[edge style={black}]{3}{4}{\color{black}dobj}
\depedge[edge style={black}]{6}{5}{\color{black}nsubj}
\depedge[edge style={black}]{1}{6}{\color{black}root}
\depedge[edge style={black}]{6}{7}{\color{black}punct}
\end{dependency}

\caption{An example dependency tree with both homogeneous (CTB) and heterogeneous (PKU) POS tags.}
\label{fig:example-dep}
\end{figure}
Among different NLP tasks, syntactic parsing is the first to convert sequential utterances into full tree structures.
Due to its simplicity and multi-lingual applicability, dependency parsing has attracted extensive research interest as a main-stream syntactic formalism \cite{nivre-lrec2016-UD1.0,zeman-etal-2018-conll}, and been widely used in semantic parsing \cite{hajic-etal-2009-conll}, information extraction \cite{roller-etal-2018-hearst}, machine translation \cite{
zhang-etal-2019-syntax}, etc.


Given an input sentence $S=w_0w_1 \ldots w_n$,
dependency parsing constructs a tree
$T=\{(h,d,l), 0\le h \le n, 1 \le d \le n, l \in \mathcal{L}\}$,
as depicted in Figure \ref{fig:example-dep},
where $(h,d,l)$ is a dependency from the head $w_h$ to the dependent $w_d$ with the relation label $l$, 
and $w_0$ is a pseudo root node.


In the pre deep learning (DL) era, part-of-speech (POS) tags are considered as indispensable ingredients for feature engineering in dependency parsing.
POS tags function as word classes in the sense that the same POS tags usually play similar syntactic roles in language utterances.
For example, ordinary verbs are tagged as \texttt{VV} and ordinary nouns as \texttt{NN} in Penn Chinese Treebank (CTB).
Yet some tags are designed for serving other tasks such as information extraction.
For instance, proper nouns and temporal nouns are distinguished from \texttt{NN} as \texttt{NR} and \texttt{NT} respectively.
In this sense, we refer to tag pairs like \{\texttt{NN},\texttt{VV}\} as syntax-sensitive and \{\texttt{NN},\texttt{NR}\} as syntax-insensitive.
Parsing performance drops dramatically when removing POS-related features, since
POS tags play a key role in reducing the data sparseness problem of using pure word-based lexical features.
Meanwhile, to alleviate error propagation in the first-tagging-then-parsing pipeline,
researchers propose to jointly model POS tagging and dependency parsing under both graph-based \cite{li-etal-2011-joint} and transition-based \cite{hatori-etal-2011-incremental} frameworks. 


In the past five years, dependency parsing has achieved tremendous progress thanks to the strong capability of deep neural networks in representing word and long-range contexts \cite{chen2014fast,dyer-etal-2015-transition,zhou-etal-2015-neural,andor-etal-2016-global,kiperwasser-goldberg-2016-simple,Timothy-d17-biaffine}.
Yet all those works hold the assumption of POS tags being important and concatenate word and POS tag embeddings as input.


Moreover, researchers show that using Character-level Long Short-term Memory (\textsc{CharLSTM}) based word representations is helpful for named entity recognition \cite{lample-etal-2016-neural}, dependency parsing \cite{dozat-etal-2017-stanfords}, 
and constituency parsing \cite{kitaev-klein-2018-constituency}. 
The idea is first to perform Long Short-term Memory (LSTM) over word characters and then to add (or concatenate) together word embeddings and \textsc{CharLSTM} word representations as model inputs.
In particular, \citet{kitaev-klein-2018-constituency} show that with \textsc{CharLSTM} word representations, POS tags are useless for constituency parsing.
We believe the reason may be two-fold.
First, word embeddings, unlike lexical features, suffer from much less data sparseness, since syntactically similar words can be associated via similar dense vectors.
Second, \textsc{CharLSTM} word representation can effectively capture morphological inflections by looking at lemma/prefix/suffix, which provide similar information as POS tags.
However, there still lacks a full and systematic study on the usefulness of POS tags for dependency parsing.

In this work, we try to answer the question whether POS tagging is necessary or even helpful for neural dependency parsing
and make the following contributions\footnote{We release our code at \url{https://github.com/Jacob-Zhou/stack-parser}.}.
\begin{itemize}
\item[$\bullet$] We design three typical multi-task learning (MTL) frameworks (i.e., \textit{Share-Loose}, \textit{Share-Tight}, \textit{Stack}), for joint POS tagging and dependency parsing based on the state-of-the-art biaffine parser.
\item[$\bullet$] Considering that there exist large-scale heterogeneous POS-tag data for Chinese partly because it is much cheaper to annotate POS tags than parse trees, we also investigate the helpfulness of such heterogeneous data, besides homogeneous POS-tag data that are annotated together with parse trees.
\item[$\bullet$] We conduct experiments on both English and Chinese benchmark datasets, and the results show that POS tagging, both homogeneous and heterogeneous, can still significantly improve parsing accuracy when using the \textit{Stack} joint framework.
Detailed analysis sheds light on the reasons behind helpfulness of POS tagging.
\end{itemize}











\section{Basic Tagging and Parsing Models}

This section presents the basic POS tagging and dependency parsing models separately in a pipeline architecture.
In order to make fair comparison with the joint models, we make the encoder-decoder architectures of the tagging and parsing models as similar as possible.
The input representation contains both word embeddings and \textsc{CharLSTM} word representations.
The encoder part adopts three BiLSTM layers.

\subsection{The Encoder Part}\label{sec:encoder-part}
\paragraph{The input layer.}
Given a sentence $S=w_0w_1 \ldots w_n$, the input layer maps each word $w_i$ into a dense vector $\mathbf{x}_i$.
\begin{equation}
\label{eq:input}
\mathbf{x}_i=\mathbf{e}_i^{w} \oplus \mathbf{e}_i^{c}
\end{equation}
where $\mathbf{e}_i^{w}$ is the word embedding, $\mathbf{e}_i^{c}$ is the \textsc{CharLSTM} word representation vector, and $\oplus$ means vector concatenation.\footnote{We have also tried the sum of $\mathbf{e}_i^{w}$ and $\mathbf{e}_i^{c}$, leading to slightly inferior performance.}
\textsc{CharLSTM} word representations $\mathbf{e}_i^{c}$ are obtained by applying BiLSTM to the word characters and concatenating the final hidden output vectors.
Following \citet{Timothy-d17-biaffine}, the word embeddings $\mathbf{e}_i^{w}$ is the sum of a fixed pretrained word embedding 
and a trainable word embedding initialized as zero.
Infrequent words in the training data (less than 2 times) are treated as a special OOV-token to learn its embedding.







Under the pipeline framework, the parsing model may use extra POS tags as input.
\begin{equation}
\mathbf{x}_i=\mathbf{e}_{i}^{w} \oplus \mathbf{e}_{i}^{c} \oplus \mathbf{e}_{i}^{p} \oplus \mathbf{e}_{i}^{p\prime}
\end{equation}
where $\mathbf{e}_{i}^{p}$ and $\mathbf{e}_{i}^{p\prime}$ are the embeddings of the homogeneous and heterogeneous POS tags, respectively.
For dropouts, we follow \citet{Timothy-d17-biaffine} and drop the different components of the input vector $\mathbf{x}_i$ independently.

\paragraph{The BiLSTM encoder.} We employ the same $N=3$ BiLSTM layers over the input layer to obtain context-aware word representations for both tagging and parsing.
We follow the dropout strategy of \citet{Timothy-d17-biaffine} and share the same dropout masks at all time steps of the same unidirectional LSTM.
The hidden outputs of the top-layer BiLSTM are used as the encoded word representations, denoted as $\mathbf{h}_i$.

\subsection{The Tagging Decoder}
\label{section:tagging-decoder}

For the POS tagging task, we use two MLP layers to compute the score vector for different tags and
get the optimal tag via softmax.
The first MLP layer uses leaky ReLU \cite{maas-2013-rectifier} activation, while the second MLP layer is linear without activation.
During training, we take the local cross-entropy loss.


\subsection{The Parsing Decoder}
\label{section:parsing-decoder}


We adopt the state-of-the-art biaffine parser of \citet{Timothy-d17-biaffine}.
We apply an MLP layer with leaky ReLU activation to obtain the representations of each word as a head ($\mathbf{r}_i^h$) and as a dependent ($\mathbf{r}_i^d$).
\begin{equation}
\mathbf{r}_i^h; \mathbf{r}_i^d =\mathrm{MLP} \left(\mathbf{h}_i \right) 
\end{equation}
As discussed in \citet{Timothy-d17-biaffine}, this MLP layer on the one hand reduces the dimensionality of $\mathbf{h}_i$, and more importantly on the other hand strips away syntax-unrelated information and thus avoids the risk of over-fitting.

Then a biaffine layer is used to 
compute scores of all dependencies.
\begin{equation}
\mathrm{score}\left(i \gets j\right)=\left[
                \begin{array}{c}
                \mathbf{r}_{i}^d \\
                1
                \end{array}
              \right]^\mathrm{T}{W} \mathbf{r}_j^h
\end{equation}
where $\mathrm{score}\left(i \gets j\right)$ is the score of the dependency $i \gets j$, and ${W}$ is a weight matrix.
During training, supposing the gold-standard head of $w_i$ is $w_j$, we use the cross-entropy loss to maximize the probability of $w_j$ being the head against all words, i.e.,
$\frac{e^{\mathrm{score}\left(i \leftarrow j \right)}}
{\sum_{0 \le k \le n} e^{\mathrm{score}\left(i \leftarrow k \right)}}$.

For dependency labels, we use extra MLP and Biaffine layers to compute the scores and also adopt cross-entropy classification loss.
We omit the details due to space limitation. 

\section{Joint Tagging and Parsing Models}

\begin{figure*}[tp]
\begin{widepage}
\centering
\includegraphics[scale=0.7]{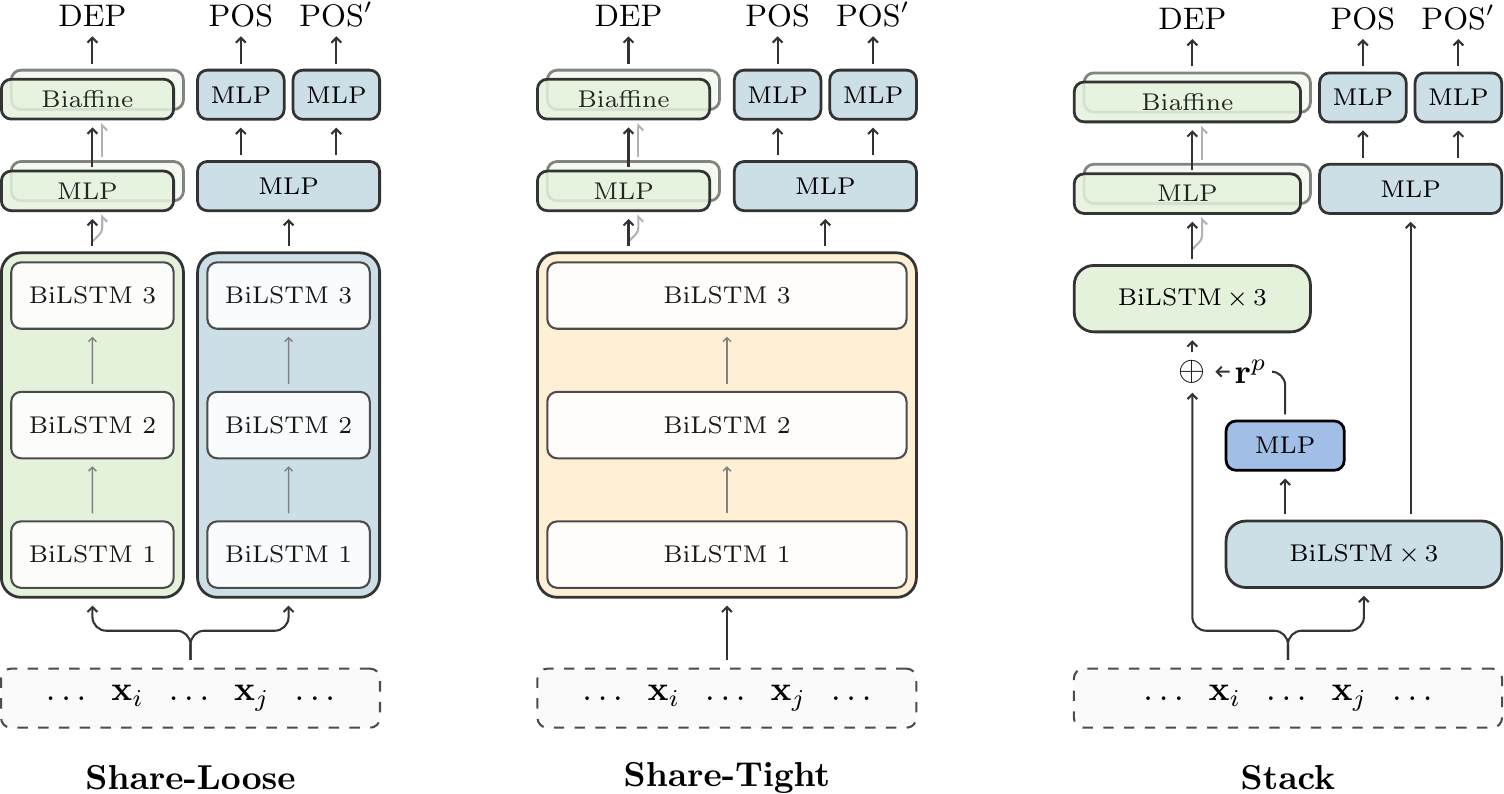}
\end{widepage}
\caption{The framework of three variants of the joint model.}
\label{fig:framework}
\end{figure*}



The pipeline framework suffers from the error propagation problem, meaning that POS tagging mistakes badly influence parsing performance.
In the pre-DL era, researchers propose joint tagging and parsing models under both graph-based and transition-based parsing architectures \cite{li-etal-2011-joint,hatori-etal-2011-incremental}.
The key idea is to define the joint score of a tag sequence and a parse tree and to find the optimal joint result in the enlarged search space.
In the neural network era, jointly modeling two tasks becomes much easier thanks to the commonly used encoder-decoder architecture and the MTL framework.

In this work, we design and compare three typical MTL frameworks for joint POS tagging and dependency parsing, i.e., \textit{Share-Loose}, \textit{Share-Tight}, and \textit{Stack}, as illustrated in Figure \ref{fig:framework}.
The \textit{Share-Loose} and \textit{Share-Tight} methods treat tagging and parsing as two parallel tasks, whereas the \textit{Stack} method consider parsing as the main task and derive POS tag-related information as the inputs of the parsing component. For all joint models, the inputs only include the word embeddings and \textsc{CharLSTM} word representations, as shown in Equation \ref{eq:input}.

\paragraph{Share-Loose.}
The tagging and parsing tasks use nearly separate networks, and only share the word and char embeddings.
To incorporate heterogeneous POS tagging data, we add another scoring MLP at the top to compute scores of different heterogeneous POS tags.
Under such architecture, the loosely connected tagging and parsing components can only influence each other in very limited manner.

\paragraph{Share-Tight.} This is the most commonly used MTL framework, in which the tagging and parsing components share not only the embeddings, but also the BiLSTM encoder.
Different decoders are then used for different tasks.
In this tightly joint model, the tagging and parsing components can interact with and mutually help each other to a large extent.
The shared parameters are trained to capture the commonalities of the two tasks.

\paragraph{Stack.}
The \textit{Stack} takes BiLSTM hidden outputs of the tagger, denoted as $\mathbf{r}_i^p$, as the extra input of the parser.
In this way, the error propagation problem can be better handled.
\begin{equation}
\mathbf{x}^{parse}_i = \mathbf{x}_i \oplus \mathbf{r}_i^p
\end{equation}
The idea of the \textit{Stack} joint method is mainly borrowed from \citet{zhang-weiss-2016-stack}.
They propose the \emph{stack-propagation} approach to avoid using explicit POS tags in dependency parsers.
They employ the simple feed-forward networks for both tagging and parsing \cite{chen2014fast}.
Without BiLSTM encoders, they use the hidden outputs of a single-layer MLP of the tagging component as extra inputs of the parsing component.

\paragraph{Training loss.} During training, we directly add together all losses of different tasks, i.e., the parsing loss, the homogeneous POS tagging loss, and the heterogeneous POS tagging loss. 
\begin{equation}
\mathcal{L}=\mathcal{L}_{\mathrm{DEP}}+\mathcal{L}_{\mathrm{POS}}+\mathcal{L}_{\mathrm{POS^\prime}}
\end{equation}
\section{Experiments}

In this section, we conduct experiments and detailed analysis to make full investigation on the usefulness of POS tagging for dependency parsing.

\subsection{Experimental Settings}

\paragraph{Data.} We conduct experiments on the English Penn Treebank (PTB), the Chinese dataset at the CoNLL-2009 shared task (CoNLL09) \cite{hajic-etal-2009-conll}, and the larger-scale Chinese Penn Treebank 7 (CTB7). 
For PTB, we adopt the same settings such as data split and Stanford dependencies of \citet{chen2014fast}.
We follow the official settings for CoNLL09.

We use the Stanford Parser v3.0 to obtain Stanford dependencies for CTB7.\footnote{\url{https://nlp.stanford.edu/software/stanford-dependencies.shtml}}
For Chinese, besides the homogeneous POS tags, we also incorporate the large-scale People Daily corpus of Peking University (PKU) as heterogeneous POS tagging data.

\paragraph{Evaluation metrics.} We use POS tagging accuracy (TA), unlabeled attachment score (UAS), and labeled attachment scores (LAS) for dependency parsing.
For UAS and LAS computation, We follow \citet{Timothy-d17-biaffine} and ignore all punctuation marks for PTB.

\paragraph{Hyper-parameters.} We follow most hyper-parameter settings of \citet{Timothy-d17-biaffine} for all our models.
For \textsc{CharLSTM} word representations, we set the dimension of the character embeddings to 50, and the dimension of \textsc{CharLSTM} outputs to 100. 
We train each model for at most 1,000 iterations, and stop training if the peak performance on the dev data does not increase in 100 (50 for models with BERT) consecutive iterations.

\subsection{Results on the Dev Data}

\paragraph{Results of the pipeline framework.}
Table \ref{table:tag-dev} shows the influence of using homogeneous and heterogeneous POS tags in the pipeline framework.
More results are also presented to understand the contributions of each of the four components in the input layer.
The homogeneous tagging accuracy is 97.58, 96.59, 96.72, 97.85, on the dev data of PTB, CoNLL09, and CTB7, and PKU, respectively.
We perform 5-fold jack-knifing to obtain the automatic homogeneous POS tags on the training data to avoid closed testing, and use a POS tagger trained on PKU data to produce heterogeneous POS tags for sentences of CoNLL09 and CTB7.

The results of using only one component clearly show that lexical information (i.e., $\mathbf{e}^w$ and $\mathbf{e}^c$) is most crucial for parsing, and only using POS tag embeddings leads to very large accuracy drop.

When using two components at the same time, using \textsc{CharLSTM} word representations ($\mathbf{e}^c$) is slightly yet consistently better than using POS tag embeddings ($\mathbf{e}^p$), both substantially outperforming the model using only word embeddings ($\mathbf{e}^w$) by more than 0.5 on all three datasets.

\begin{semiwidepage}
    \begin{minipage}[tb]{0.47\textwidth}
        \centering
        \setlength{\tabcolsep}{1.5pt}
\begin{tabular}{lcccccc}
\toprule
& PTB & CoNLL09 & CTB7 \\\\[-12pt]
\hline\\[-10pt]
$\mathbf{e}^p$ & 87.79 & 75.94 & 75.72 \\
$\mathbf{e}^w$ & 93.42 & 85.30 & 84.43 \\
$\mathbf{e}^c$ & 93.34 & 84.42 & 83.73 \\\\[-12pt]
\hline\\[-10pt]

$\mathbf{e}^w \oplus \mathbf{e}^p$ & 93.92 & 85.94 & 85.12 \\

$\mathbf{e}^w \oplus \mathbf{e}^c$ & \textbf{93.97} & 86.09 & 85.23 \\\\[-12pt]
\hline\\[-10pt]
$\mathbf{e}^w \oplus \mathbf{e}^c \oplus \mathbf{e}^p$ & 93.88 & \textbf{86.17} & \textbf{85.32} \\
$\mathbf{e}^w \oplus \mathbf{e}^c \oplus \mathbf{e}^p \oplus \mathbf{e}^{p\prime}$ & - & 86.01 & 85.23 \\
\bottomrule
\end{tabular}


        \captionof{table}{Parsing performance (LAS) on dev data under the pipeline framework.}
        \label{table:tag-dev}
    \end{minipage}
    \hfill
    \begin{minipage}[tb]{0.47\textwidth}
        \centering
        \setlength{\tabcolsep}{1.5pt}
\begin{tabular}{llccc}
\toprule
& & PTB & CoNLL09 & CTB7 \\\\[-12pt]
\hline\\[-10pt]
\multirow{3}{*}{\shortstack{homo}}
& \textit{Share-Loose}    &         93.95  &         86.28  &         85.56  \\
& \textit{Share-Tight}    &         93.93  &         86.17  &         85.56  \\
& \textit{Stack}          & \textbf{94.09} & \textbf{86.26} & \textbf{85.79} \\\\[-9pt]
\hline\\[-7pt]
\multirow{3}{*}{\shortstack{hetero}}
& \textit{Share-Loose}    & - &         86.05  &         85.62  \\
& \textit{Share-Tight}    & - & \textbf{86.25} &         85.76  \\
& \textit{Stack}          & - &         86.16  & \textbf{85.86} \\\\[-9pt]
\hline\\[-7pt]
\multirow{3}{*}{\shortstack{homo\\+\\hetero}}
& \textit{Share-Loose}    & - &         86.30  &         85.57  \\
& \textit{Share-Tight}    & - &         86.62  &         85.86  \\
& \textit{Stack}          & - & \textbf{86.69} & \textbf{85.88} \\
\bottomrule
\end{tabular}
        \captionof{table}{Parsing performance (LAS) comparison on dev data for the three joint methods.}
        \label{table:dev}
    \end{minipage}
    \vspace{0.5cm}
\end{semiwidepage}

Moreover, using three components leads to slight improvement on both CoNLL09 and CTB7, but hurts performance on PTB.
Further using heterogeneous tag embeddings slightly degrades the performance.

All those results indicate that under the pipeline framework, POS tags become unnecessary and can be well replaced by the \textsc{CharLSTM} word representations.
We believe the reasons are two-fold.
First, \textsc{CharLSTM} can effectively capture morphological inflections by looking at lemma/prefix/suffix, and thus plays a similar role as POS tags in terms of
alleviating the data sparseness problem of words.
Second, the error propagation issue makes predicted POS tags less reliable.



\paragraph{Results of the joint methods.}
Table \ref{table:dev} presents the results of the three joint tagging and parsing methods without or with heterogeneous POS tagging.

When using only homogeneous POS tagging, we find that the performance gaps between different joint methods are very small.
The best joint methods outperform the basic model by 0.1, 0.3, and 0.6 respectively.
A similar situation arises when using heterogeneous tagging only.

When using both homogeneous and heterogeneous tagging , aka (w/ hetero) setting, we can see that the overall performance is further improved by large margin.
The best \textit{Stack} method outperforms the basic model by 0.6 on CoNLL09 and 0.7 on CTB7, showing that heterogeneous labeled data can inject useful knowledge into the model.

Overall, we can see that the \textit{Stack} method is more stable and superior compared with the other three methods, and is adopted for the following experiments and analysis.

\subsection{Final Results on the Test Data}

\begin{table}[tb]
\setlength{\tabcolsep}{1.5pt}
\begin{widepage}
\centering
\begin{tabular}{llccccccccc}
\toprule
& & \multicolumn{3}{c}{PTB} & \multicolumn{3}{c}{CoNLL09} & \multicolumn{3}{c}{CTB7} \\
& & TA & UAS & LAS & TA & UAS & LAS & TA & UAS & LAS \\\\[-12pt]
\hline\\[-10pt]
\multirow{7}{*}{w/o BERT}
& \citet{andor-etal-2016-global} & 97.44 & 94.61 & 92.79 & - & 84.72 & 80.85 & - & - & - \\
& \citet{Timothy-d17-biaffine} & 97.3\,\,\,  & 95.74 & 94.08 & - & 88.90 & 85.38 & - & - & - \\
& \citet{ji-etal-2019-graph} & 97.3\,\,\,  & 95.97 & 94.31 & - & - & - & - & - & - \\
& \citet{li-etal-2019-attentive} & 97.3\,\,\,  & 95.93 & 94.19 & - & 88.77 & 85.58 & - & - & - \\
& Basic ($\mathbf{e}^w \oplus \mathbf{e}^c$) & 97.50 & 95.97 & 94.34 & 96.42 & 89.12 & 86.00 & 96.48 & 88.58 & 85.40 \\
& Pipeline ($\mathbf{e}^w \oplus \mathbf{e}^c \oplus \mathbf{e}^p$) & 97.50 & 95.88 & 94.27 & 96.42 & 89.12 & 85.98 & 96.48 & 88.42 & 85.28 \\
& \textit{Stack} & \textbf{97.91} & \textbf{96.13} & \textbf{94.53} & 96.55 & 89.46 & 86.44 & 96.62 & 88.86 & 85.88 \\
& \textit{Stack} w/ hetero & - & - & - & \textbf{96.66} & \textbf{89.85} & \textbf{86.85} & \textbf{96.72} & \textbf{89.26} & \textbf{86.27} \\\\[-12pt]
\hline\\[-10pt]
\multirow{4}{*}{w/ BERT}
& \citet{li-etal-2019-attentive} & - & 96.67 & 95.03 & - & 92.24 & 89.29 & - & - & - \\
& Basic ($\mathbf{e}^w \oplus \mathbf{e}^c$) & 97.42 & 96.85 & 95.14 & 97.29 & 92.21 & 89.42 & 97.22 & 91.66 & 88.75 \\
& \textit{Stack}  & \textbf{97.57} & \textbf{96.85} & \textbf{95.25} & 97.36 & 92.44 & 89.68 & 97.32 & 91.67 & 88.84 \\
& \textit{Stack} w/ hetero & - & - & - & \textbf{97.39} & \textbf{92.46} & \textbf{89.76} & \textbf{97.40} & \textbf{91.81} & \textbf{89.04} \\
\bottomrule
\end{tabular}
\end{widepage}
\caption{Final results on the test data. It is noteworthy that we produce our experiments with single run for each model on each dataset, since our preliminary experiments that we train \textit{Stack} w/ hetero and Basic on CTB7 for four times show the variance of performances is small ($\sigma^2 < 0.01$). }
\label{table:final-results}
\end{table}

Table \ref{table:final-results} shows the results on the test data.
For the scenario of not using BERT, the pipeline method using homogeneous POS tags is slightly yet consistently inferior to the basic model.
The \textit{Stack} method using only homogeneous POS tags significantly outperforms the basic method by 0.2 ($p<0.005$), 0.4 ($p<0.0005$), and 0.5 ($p<0.0005$) in LAS on the three datasets respectively.
Utilizing heterogeneous POS tags on Chinese further \textcolor{black}{boosts} parsing performance, leading to large overall improvements of 0.9 ($p<0.0001$) on both datasets.

When using BERT, parsing accuracy of the basic method increases by very large margin.
Compared with the stronger baseline, the improvement introduced by POS tagging becomes smaller.
Overall, using both homogeneous and heterogeneous POS tagging, the \textit{Stack} method significantly outperforms the basic method by 0.3 ($p<0.005$) on both CoNLL09 and CTB7.



For POS tagging, the trend of performance change is similar.
First, the joint method can also improve tagging accuracy, especially when with heterogeneous POS tagging.
Using Bert can substantially improve TA on both Chinese datasets.
However, it is surprising to see a slight decrease in TA when using BERT, which is possibly due to over-fitting considering the TA is already very high on English.

We also list the results of recent previous works.
We can see that our final joint models achieve competitive parsing accuracy on PTB and CoNLL09 w/ or w/o BERT. 

\subsection{Detailed Analysis}
In the following, we conduct detailed analysis on the CoNLL09 test data, in order to understand or gain more insights on the interactions and mutual influence between POS tagging and dependency parsing.
For the joint method, we adopt the \textit{Stack} model with both homogeneous and heterogeneous POS tagging without using BERT, to jointly produce automatic POS tags and parse trees.
For the pipeline method, we use the two basic tagging and parsing models separately to produce automatic results.
\begin{figure}[tb]
\centering
\includegraphics{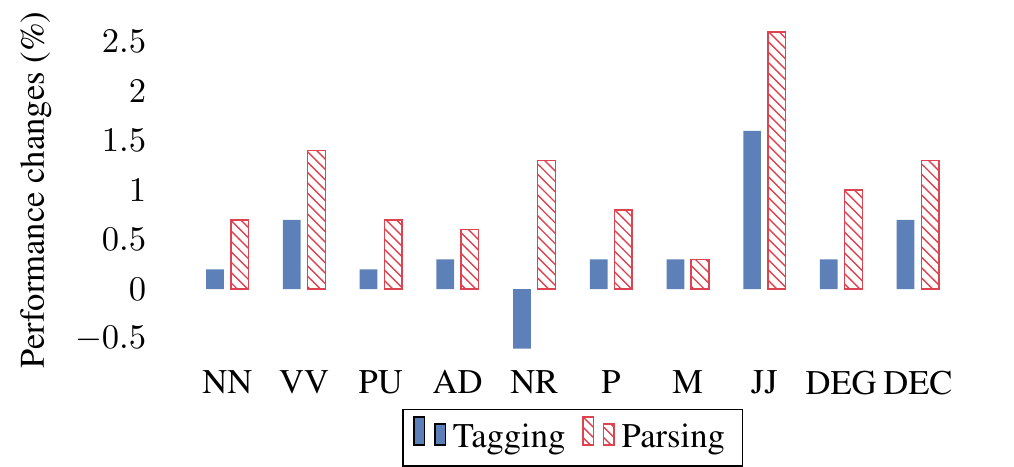}
\caption{Changes of tagging accuracy and parsing accuracy (LAS) on the CoNLL09 test set for words of different POS tags. Words/arcs are categorized according to their/their dependent's gold-standard POS tags.}
\label{fig:tp-bar}
\end{figure}

\paragraph{Correlation of performance changes between tagging and parsing.}
Overall, the joint method outperforms the pipeline method by 0.2 in TA, and 0.7/0.9 in UAS/LAS, as shown in Table \ref{table:final-results}.
To gain more insights, we categorize all words according to their gold-standard POS tags and compare the accuracy changes for each set.
Figure \ref{fig:tp-bar} shows the most frequent tags. 
We can see that there is clear positive correlation of absolute performance changes between tagging and parsing.
For instance, as the most frequent tags \texttt{NN} and \texttt{VV}, their tagging accuracy increases by 0.2 and 0.7, and parsing accuracy increases by 0.7 and 1.4, respectively.
The most notable exception is \texttt{NR} with opposite changes in tagging and parsing accuracy (-0.6 vs. +1.3), which can be explained from two aspects.
First, we find that most of \texttt{NR} mistakes are due to the \{\texttt{NR}, \texttt{NN}\} ambiguous pair, which is syntax-insensitive and thus has very small impact on parsing decisions.
Second, the \textit{Stack} model may be more robust to tagging errors.

Overall, we conclude that tagging and parsing performance is highly correlated due to the close relationship between the two tasks.

\paragraph{Influence of tagging errors on parsing.}
In the \textit{Stack} method, the hidden representations from the tagging encoder is fed into the parsing encoder as extra inputs.
We would like to understand how tagging decisions influence parsing.
Overall, UAS/LAS are 90.42/88.38 for words getting correct POS tags, whereas 73.40/42.59 for wrongly tagged words.
We can observe dramatic drop of 17.0/45.8, indicating that POS tags has much larger influence on LAS than UAS.
\begin{table}[tb]
\centering
\begin{tabular}{rlcc||rlcc}
\toprule
& & UAS & LAS & & & UAS & LAS \\\\[-12pt]
\hline\\[-10pt]
\texttt{NN} & $\rightarrow$ \texttt{NN} & 91.73 & 89.69 & \texttt{NR}  & $\rightarrow$ \texttt{NR}  & 91.73 & 86.96 \\
            & $\rightarrow$ \texttt{VV} & 67.25 & 44.98 &              & $\rightarrow$ \texttt{NN}  & 86.39 & 83.67 \\
            & $\rightarrow$ \texttt{NR} & 90.43 & 86.96 & \texttt{JJ}  & $\rightarrow$ \texttt{JJ}  & 95.40 & 94.33 \\
            & $\rightarrow$ \texttt{JJ} & 91.96 & 20.54 &              & $\rightarrow$ \texttt{NN}  & 92.82 & 14.92 \\
\texttt{VV} & $\rightarrow$ \texttt{VV} & 85.92 & 84.12 & \texttt{DEG} & $\rightarrow$ \texttt{DEG} & 96.75 & 95.91 \\
            & $\rightarrow$ \texttt{NN} & 65.60 & 40.07 &              & $\rightarrow$ \texttt{DEC} & 92.06 & 26.56 \\
            & $\rightarrow$ \texttt{VA} & 84.75 & 83.05 & \texttt{DEC} & $\rightarrow$ \texttt{DEC} & 94.28 & 92.39 \\
            & $\rightarrow$ \texttt{AD} & 55.32 & 25.53 &              & $\rightarrow$ \texttt{DEG} & 96.88 & 22.22 \\
\bottomrule
\end{tabular}
\caption{The impact of specific POS tagging error patterns on parsing.}
\label{table:tag-pair}
\end{table}

Looking deeper into this issue, Table \ref{table:tag-pair} shows the parsing accuracy for words of different POS tagging patterns.
A tagging pattern \texttt{X} $\rightarrow$ \texttt{Y} represents the set of words whose correct tag is \texttt{X} and \textcolor{black}{are} tagged as \texttt{Y}.

We can see that higher parsing accuracy are usually achieved by correct tagging patterns \texttt{X} $\rightarrow$ \texttt{X} than wrong pattern \texttt{X} $\rightarrow$ \texttt{NOT-X}, except 
\texttt{DEC} $\rightarrow$ \texttt{DEG} in UAS.\footnote{\texttt{DEG} and \texttt{DEC} are two tags for the frequently used auxiliary word ``的'' ($\mathrm{d\bar{e}}$, translated as ``of'' or ``that'') in Chinese. ``的'' is tagged as \texttt{DEG} in phrase ``土地/land 的 面积/area (area of the land)'', while as \texttt{DEC} in ``他/he 提出/proposed 的 方法/method (method that he proposed)''.}.

The tagging ambiguites can be classified into three types.
First, the syntax-sensitive ambiguous pairs such as \{\texttt{NN}, \texttt{VV}\} and \{\texttt{VV}, \texttt{AD}\} lead to large performance decrease in both UAS and LAS if wrongly tagged.
Second, the syntax-insensitive ambiguous pairs such as \{\texttt{NN}, \texttt{NR}\} and \{\texttt{VV}, \texttt{VA}\} have very small influence on parsing accuracy.
Finally, some ambiguous pairs only greatly influence LAS but have little effect on UAS, such as \{\texttt{NN}, \texttt{JJ}\} and \{\texttt{DEC}, \texttt{DEG}\}.

\section{Related Works}

Previous studies \cite{mcdonald-d11-delexicalized,lei-etal-2014-low} show POS tags are indispensable ingredients for composing different features in the traditional pre-DL dependency parsers \cite{koo-acl10-3o,zhang-nivre-p11-transition}.
Meanwhile, \citet{li-etal-2011-joint} show that the error propagation problem introduced by predicted POS tags degrades parsing accuracy by about 6 (UAS) on different Chinese  datasets.
Therefore, researchers propose to jointly model the POS tagging and dependency tasks in the graph-based \cite{li-etal-2011-joint} and transition-based \cite{hatori-etal-2011-incremental} 
frameworks, leading to promising results.
The key challenge is to define scoring functions on the joint results, and design effective search algorithms to determine optimal joint answers in the enlarged search space.
Furthermore, joint models of word segmentation, POS tagging, and parsing are also proposed \cite{hatori-EtAl:2012:ACL2012}.

In the DL era, joint modeling of multiple related tasks becomes much easier under the MTL framework \cite{collobert-08-multi-task}.
In fact, MTL has become an extensively used and powerful technique for many problems.
The basic idea is sharing the encoder part while using separate decoders for different tasks.
The major advantages of employing MTL are two-fold, i.e., 1) exploiting the correlation and mutual helpfulness among related tasks, and 2) making direct use of all (usually non-overlapping) labeled data of different tasks.
The Share-Light and Share-Tight methods are both typical MTL frameworks, and the main difference lies in the amount of shared parameters.
Actually, there are still many other variants due to the flexibility of MTL.
For example, \citet{straka-2018-udpipe} stacks task-specific private BiLSTMs over shared BiLSTMs for joint tagging and parsing.
Based on the current results, we expect that such variants may achieve very similar performance.


The \textit{Stack} method is similar to the \emph{stack-propagation} method of \citet{zhang-weiss-2016-stack}.
Their basic idea is to use the hidden outputs of the POS tagging components as extra inputs of the parsing components, forming a stacked structure.
During training, parsing loss is directly propagated into the full tagging component whereas tagging loss only indirectly influences the parsing components via their shared parts.
Their pioneer work employ a simple feed-forward network for both tagging and parsing \cite{chen2014fast}, and only achieves an LAS of 91.41 on PTB.   
Another inspiring work related with the \textit{Stack} method is \citet{hashimoto-etal-2017-joint}, who propose to jointly train many tasks of different complexity in a very deep and cascaded network architecture, where higher levels are used for more complex tasks.

\section{Conclusions}

Unlike the findings in traditional pre-DL dependency parsing, recent studies indicate that POS tagging becomes much less important and can be replaced by \textsc{CharLSTM} word representations in neural dependency parsers.
However, there lacks a full and systematic investigation on this interesting issue, from both empirical and linguistic perspectives.
In this paper, we try to investigate the role of POS tagging for neural dependency parsing in both pipeline and joint frameworks.
We design and compare three typical joint methods based on the state-of-the-art biaffine parser.
We try to accommodate both homogeneous and heterogeneous POS tagging, considering it is much cheaper to annotate POS tags than parse trees and there exist large-scale heterogeneous POS tag datasets for Chinese.
Based on the experiments and analysis on three English and Chinese benchmark datasets, we can draw the following conclusions.
\begin{itemize}
    \item[$\bullet$] For the pipeline method, both homogeneous and heterogeneous POS tags provide little help to the basic parser with both word embeddings and \textsc{CharLSTM}, due to error propagation and the overlapping role in reducing data sparseness.
    \item[$\bullet$] The three joint methods investigated in this work perform better than the pipeline method. Among them, the \textit{Stack} is more stable and superior compared with the other three, leading to significant improvement over the basic model on all datasets.
    \item[$\bullet$] POS tagging is still helpful for dependency parsing under the joint framework even if the parser is enhanced with BERT, especially when with heterogeneous POS tagging.
    \item[$\bullet$] Detailed analysis shows that POS tagging and dependency parsing are two closely correlated tasks. In particular, If the joint model fails to resolve syntax-sensitive POS tagging ambiguities, it usually makes wrong parsing decisions as well.
\end{itemize}



\renewcommand{\bibsection}{\section*{References}} 
\bibliographystyle{splncsnat}
\small 
\bibliography{nlpcc2020}

\end{CJK*}

\end{document}